\documentclass{article}

\usepackage{hyperref}

\usepackage[accepted]{icml2024}

\usepackage{amsmath,amsfonts,bm}

\def\eqref#1{equation~\ref{#1}}

\def\1{\bm{1}}

\def\vm{{\bm{m}}}

\def\vv{{\bm{v}}}

\DeclareMathAlphabet{\mathsfit}{\encodingdefault}{\sfdefault}{m}{sl}
\SetMathAlphabet{\mathsfit}{bold}{\encodingdefault}{\sfdefault}{bx}{n}

\DeclareMathOperator*{\argmax}{arg\,max}

\usepackage[utf8]{inputenc} %
\usepackage{lipsum} %

\usepackage[T1]{fontenc}

\usepackage{etoolbox}
\newbool{includeappendix}
\setbool{includeappendix}{true}

\ifdefined\isoverfull
\else
\fi

\usepackage{xcolor} %

\definecolor{my-full-blue}{HTML}{1F77B4}

\definecolor{my-full-orange}{HTML}{FF7F0E}

\definecolor{my-full-green}{HTML}{2CA02C}

\definecolor{my-full-red}{HTML}{d62728}

\definecolor{my-full-purple}{HTML}{9467bd}

\definecolor{my-full-yellow}{HTML}{F4D867}

\colorlet{my-blue}{my-full-blue!30}
\colorlet{my-orange}{my-full-orange!30}
\colorlet{my-green}{my-full-green!30}
\colorlet{my-red}{my-full-red!30}
\colorlet{my-purple}{my-full-purple!30}
\colorlet{my-yellow}{my-full-yellow!30}

\colorlet{my-lightgray}{lightgray!30}

\usepackage{listings}

\usepackage{textcomp}

\usepackage{xcolor}

\usepackage[scaled=0.8]{beramono}

\definecolor{ckeyword}{HTML}{7F0055}
\definecolor{ccomment}{HTML}{3F7F5F}
\definecolor{cstring}{HTML}{2A0099}

\lstdefinestyle{numbers}{
	numbers=left,
	framexleftmargin=20pt,
	numberstyle=\tiny,
	firstnumber=auto,
	numbersep=1em,
	xleftmargin=2em
}

\lstdefinestyle{layout}{
	frame=none,
	captionpos=b,
}

\lstdefinestyle{comment-style}{
	morecomment=[l]//,
	morecomment=[s]{/*}{*/},
	commentstyle={\color{ccomment}\itshape},
}

\lstdefinestyle{string-style}{
	morestring=[b]",%
	morestring=[b]',%
	stringstyle={\color{cstring}},
	showstringspaces=false,%
}

\lstdefinestyle{keyword-style}{
	keywordstyle={\ttfamily\bfseries},
	morekeywords={
		function,
		constructor,
		int,
		bool,
		return,
		returns,
		uint
	},
	morekeywords = [2]{},
	keywordstyle = [2]{\text},
	sensitive=true,
}

\lstdefinestyle{input-encoding}{
	inputencoding=utf8,
	extendedchars=true,
	literate=
	{ℝ}{$\reals$}1%
	{→}{$\rightarrow$}1%
	{α}{$\alpha$}1%
	{β}{$\beta$}1%
	{λ}{$\lambda$}1%
	{θ}{$\theta$}1%
	{ϕ}{$\phi$}1%
}

\lstdefinestyle{escaping}{
	moredelim={**[is][\color{blue}]{\%}{\%}},
	escapechar=|,
	mathescape=true
}

\lstdefinestyle{default-style}{
	basicstyle=\fontencoding{T1}\ttfamily\footnotesize,
	style=numbers,
	style=layout,
	style=comment-style,
	style=string-style,
	style=keyword-style,
	style=input-encoding,
	style=escaping,
	tabsize=2,
	upquote=true
}

\lstdefinelanguage{BASIC}{
	language=C++,
	style=default-style
}[keywords,comments,strings]%

\usepackage[capitalize]{cleveref}

\crefformat{section}{\S#2#1#3}

\crefrangeformat{section}{\S#3#1#4\crefrangeconjunction\S#5#2#6}

\crefmultiformat{section}{\S#2#1#3}{\crefpairconjunction\S#2#1#3}{\crefmiddleconjunction\S#2#1#3}{\creflastconjunction\S#2#1#3}

\newcommand{\crefrangeconjunction}{--}

\crefname{listing}{Lst.}{listings}
\crefname{line}{Lin.}{Lin.}
\crefname{appendix}{App.}{App.}

\newcommand{\app}[1]{%
	\ifbool{includeappendix}{\cref{#1}}{the appendix}%
}
\newcommand{\App}[1]{%
	\ifbool{includeappendix}{\cref{#1}}{The appendix}%
}

\usepackage{tikz}

\usetikzlibrary{arrows}
\usetikzlibrary{automata}
\usetikzlibrary{calc}
\usetikzlibrary{backgrounds}
\usetikzlibrary{decorations.markings}
\usetikzlibrary{decorations.pathmorphing}
\usetikzlibrary{decorations.pathreplacing}
\usetikzlibrary{fit}
\usetikzlibrary{patterns}
\usetikzlibrary{positioning}
\usetikzlibrary{shadows}
\usetikzlibrary{shapes}
\usetikzlibrary{shapes.geometric}

\usepackage{microtype}
\usepackage{graphicx}
\usepackage{subfigure}
\usepackage{booktabs} %
\usepackage{pifont}
\usepackage{xspace}
\usepackage{eqparbox,array}
\usepackage{adjustbox}
\usepackage{multirow}
\usepackage[flushleft]{threeparttable}
\usepackage{calc}
\usepackage{makecell}
\usepackage{setspace}
\usepackage{enumitem}
\usepackage{lipsum}
\usepackage{nicefrac}
\usepackage{tcolorbox}

\renewcommand\algorithmiccomment[1]{%
  \hfill// #1%
}
\newcommand{\alglinelabel}{%
  \addtocounter{ALC@line}{-1}%
  \refstepcounter{ALC@line}%
  \label%
}
\crefname{ALC@line}{line}{lines}
\makeatletter
\newcommand{\DEF}[1]{\ALC@it%
\textbf{def} #1: \begin{ALC@rpt}}%
\newcommand{\ENDDEF}{\end{ALC@rpt}}%
\makeatother

\newtheorem{theorem}{Theorem}[section]
\newtheorem{definition}{Definition}[section]
\newtheorem{lemma}[theorem]{Lemma}

\newlength{\DepthReference}
\newlength{\HeightReference}
\newlength{\Width}%
\newlength{\Height}%

\newcommand{\MyColorBox}[2][red]%
{%
    \settowidth{\Width}{#2}%
    \settoheight{\Height}{#2}%
    \settodepth{\Height}{#2}%
    \setlength{\fboxsep}{0pt}%
    \colorbox{#1}%
    {%
                \parbox[t][\HeightReference+\DepthReference][b]{\Width}{\centering\raisebox{\DepthReference}{#2}}%
    }%
}

\newcommand{\MyFColorBox}[2][red]%
{%
    \settowidth{\Width}{#2}%
    \settoheight{\Height}{#2}%
    \settodepth{\Height}{#2}%
    \setlength{\fboxsep}{0pt}%
    \fcolorbox{red}{#1}%
    {%
                \parbox[t][\HeightReference+\DepthReference][b]{\Width}{\centering\raisebox{\DepthReference}{#2}}%
    }%
}

\newcommand{\MyTemplateBox}[2][blue]%
{%
    \settowidth{\Width}{#2}%
    \settoheight{\Height}{#2}%
    \settodepth{\Height}{#2}%
    \setlength{\fboxsep}{0pt}%
    \fcolorbox{blue}{#1}%
    {%
                \parbox[t][\HeightReference+\DepthReference][b]{\Width}{\centering\raisebox{\DepthReference}{#2}}%
    }%
}

\newcommand{\yes}{\checkmark\xspace}
\newcommand{\yesoffline}{\checkmark\xspace}
\newcommand{\yesonline}{\checkmark\xspace}
\newcommand{\yespartial}{(\checkmark)\xspace}

\newcommand{\partialonline}{(\checkmark)\xspace}
\newcommand{\no}{\ding{55}}

\newcommand{\picard}{\textsc{Picard}\xspace}
\newcommand{\lmql}{\textsc{lmql}\xspace}
\newcommand{\synchromesh}{\textsc{Synchromesh}\xspace}
\newcommand{\outlines}{\textsc{Outlines}\xspace}
\newcommand{\guidance}{\textsc{guidance}\xspace}
\newcommand{\llamacpp}{\textsc{llama.cpp}\xspace}
\newcommand{\tgcd}{\textsc{gcd}\xspace}

\newcommand{\tool}{\textsc{Domino}\xspace}

\newcommand{\voc}{\ensuremath{\mathcal{V}}\xspace}

\newcommand{\ltok}{\ensuremath{l}\xspace}

\newcommand{\token}[3][my-lightgray]{\MyColorBox[my-full-orange!#3!#1]{\texttt{#2}}\hspace{0.8pt}}
\newcommand{\dtoken}[3][my-lightgray]{\raisebox{\HeightReference}{\token[#1]{#2}{#3}}\xspace}

\newcommand{\atoken}[3][my-lightgray]{\MyFColorBox[my-full-orange!#3!#1]{\texttt{#2}}\hspace{0.8pt}}
\newcommand{\adtoken}[3][my-lightgray]{\raisebox{\HeightReference}{\atoken[#1]{#2}{#3}}\xspace}

\newcommand{\btoken}[3][my-lightgray]{\MyTemplateBox[my-full-blue!#3!#1]{\texttt{#2}}\hspace{0.8pt}}
\newcommand{\abtoken}[3][my-lightgray]{\raisebox{\HeightReference}{\btoken[#1]{#2}{#3}}\xspace}

\newcommand{\tcursor}[3][white]{\vspace{0.2em}\hspace{-0.05em}\raisebox{\HeightReference}{\token[#1]{#2}{#3}}}
\newcommand{\ttoken}[2][my-lightgray]{\raisebox{\HeightReference}{\token[#1]{#2}{0}}\xspace}

\colorlet{accept}{my-full-green!30}
\colorlet{reject}{my-full-red!30}
\colorlet{undecided}{my-full-yellow!50}

\newcommand{\ret}{\textbackslash{}n}
\newcommand{\eos}{\textsc{EOS}}
\newcommand{\tabchar}{\textbackslash{}t}

\renewcommand{\algorithmicrequire}{\textbf{Input:}}
\renewcommand{\algorithmicensure}{\textbf{Output:}}
\newcommand{\ttt}[1]{\text{\texttt{#1}}}

\newcommand{\terminalstart}[1]{\tikz[]{
   \raisebox{-0.25mm}{
      \node[rectangle, fill=#1, inner sep=0pt, minimum height=2.5mm, minimum width=2mm] (node) {};
      \draw[draw=black]
                  ($(node.south east)+(0mm,0mm)$) --
                  ($(node.south west)+(0mm,0mm)$);
      \draw[draw=black]
                  ($(node.north west)+(0mm,0mm)$) --
                  ($(node.south west)+(0mm,0mm)$);
      \draw[draw=black]
                  ($(node.north west)+(0mm,0mm)$) --
                  ($(node.north east)+(0mm,0mm)$);
  }
}\xspace}

\newcommand{\terminalstartme}[1]{\tikz[]{
   \raisebox{-0.25mm}{
      \node[rectangle, fill=#1, inner sep=0pt, minimum height=2.5mm, minimum width=2mm] (node) {};
      \draw[draw=black]
                  ($(node.south east)+(0mm,0mm)$) --
                  ($(node.south west)+(0mm,0mm)$);
      \draw[draw=black]
                  ($(node.north west)+(0mm,0mm)$) --
                  ($(node.south west)+(0mm,0mm)$);
      \draw[draw=black]
                  ($(node.north west)+(0mm,0mm)$) --
                  ($(node.north east)+(0mm,0mm)$);
      \draw[draw=black, densely dotted]
                  ($(node.north east)+(0mm,0mm)$) --
                  ($(node.south east)+(0mm,0mm)$);

  }
}\xspace}

 \newcommand{\terminalmiddle}[1]{
  \tikz[]{
   \raisebox{-0.25mm}{
      \node[rectangle, fill=#1, inner sep=0pt, minimum height=2.5mm, minimum width=2mm] (node) {};
      \draw[draw=black]
                  ($(node.south west)+(0mm,0mm)$) --
                  ($(node.south east)+(0mm,0mm)$);
      \draw[draw=black]
                  ($(node.north east)+(0mm,0mm)$) -- 
                  ($(node.north west)+(0mm,0mm)$);
  }
 }\xspace}
 
 \newcommand{\terminalmiddleme}[1]{
  \tikz[]{
   \raisebox{-0.25mm}{
      \node[rectangle, fill=#1, inner sep=0pt, minimum height=2.5mm, minimum width=2mm] (node) {};
      \draw[draw=black]
                  ($(node.south west)+(0mm,0mm)$) --
                  ($(node.south east)+(0mm,0mm)$);
      \draw[draw=black]
                  ($(node.north east)+(0mm,0mm)$) -- 
                  ($(node.north west)+(0mm,0mm)$);
      \draw[draw=black, densely dotted]
                  ($(node.north east)+(0mm,0mm)$) --
                  ($(node.south east)+(0mm,0mm)$);

  }
 }\xspace}

 \newcommand{\terminalend}[1]{
   \raisebox{-0.25mm}{
   \tikz[]{
      \node[rectangle, fill=#1, inner sep=0pt, minimum height=2.5mm, minimum width=2mm] (node) {};
      \draw[draw=black]
                  ($(node.south west)+(0mm,0mm)$) --
                  ($(node.south east)+(0mm,0mm)$);
      \draw[draw=black]
                  ($(node.north east)+(0mm,0mm)$) -- 
                  ($(node.north west)+(0mm,0mm)$);
       \draw[draw=black]
                  ($(node.south east)+(0mm,0mm)$) --
                  ($(node.north east)+(0mm,0mm)$);
  }
  }
  \hspace{-2.5mm}
  \xspace}

  \newcommand{\terminal}[1]{\tikz[]{
   \raisebox{-0.25mm}{
      \node[rectangle, fill=#1, inner sep=0pt, minimum height=2.5mm, minimum width=2mm] (node) {};
      \draw[draw=black]
                  ($(node.south west)+(0mm,0mm)$) --
                  ($(node.south east)+(0mm,0mm)$) --
                  ($(node.north east)+(0mm,0mm)$) -- 
                  ($(node.north west)+(0mm,0mm)$) --
                  ($(node.south west)+(0mm,0mm)$);
  }
  }\xspace}

 \newcommand{\bgbox}[2]{
 \fcolorbox{white}{my-lightgray!50}{%
 \begin{minipage}{\dimexpr\textwidth-2\fboxrule-2\fboxsep\relax}
  {\scriptsize \textbf{#1\vfill}\hspace{0.3em}}
  \centering
   #2
 \end{minipage}}
 }

\def\titlevar{Guiding LLMs The Right Way: Fast, Non-Invasive Constrained Generation}
\icmltitlerunning{\titlevar}

\begin{document}

\twocolumn[
\icmltitle{\titlevar}

\icmlsetsymbol{equal}{*}

\begin{icmlauthorlist}
\icmlauthor{Luca Beurer-Kellner}{eth}
\icmlauthor{Marc Fischer}{eth}
\icmlauthor{Martin Vechev}{eth}
\end{icmlauthorlist}

\icmlaffiliation{eth}{Department of Computer Science, ETH Zurich, Switzerland}

\icmlcorrespondingauthor{Luca Beurer-Kellner}{
	luca.beurer-kellner@inf.ethz.ch}
\icmlcorrespondingauthor{Marc Fischer}{marc.fischer@inf.ethz.ch}

\icmlkeywords{Machine Learning, ICML}

\vskip 0.3in
]

\printAffiliationsAndNotice{\icmlEqualContribution} %

\begin{abstract}
To ensure that text generated by large language models (LLMs) is in an expected format, constrained decoding proposes to enforce strict formal language constraints during generation. However, as we show in this work, not only do such methods incur performance overhead during generation, but many of them also significantly impair task accuracy, if they do not correctly align the underlying LLM sub-word vocabularies with external constraints. To address this, we present a novel decoding algorithm, DOMINO, that can enforce constraints in a fully subword-aligned fashion, while leveraging pre-computation and speculative decoding to achieve virtually no overhead and in some cases even almost 2$\times$ speedup over unconstrained decoding -- thereby outperforming existing approaches by a wide margin.

\end{abstract}

\section{Introduction} \label{sec:introduction}

The recent success of Large Language Models (LLMs) \citep{BrownMRSKDSKSSA20,ChenTJYPKEBP21,gpt4,TouvronLIMLLRAGJGL23,TouvronMSAAB23,Gemini,JiangRSBLSBCLSSSALBGLLSSYATLWLE23} has lead to the development of various methods that facilitate constrained generation, a method that lets users tailor the output of an LLM to a specific task or format.

\paragraph{Constrained Decoding}
To ensure that text generated by an LLM adheres to syntactic constraints, these methods restrict the decoding procedure of an LLM in a way that only permits syntactically valid tokens at each sampling step. Doing so, the generated text can be ensured to adhere to constraints like high-level templates \citep{Beurer-Kellner023,guidance}, regular expressions \citep{Beurer-Kellner023,guidance,WillardL23} or context-free grammars \citep{WillardL23,guidance}.
Constrained decoding provides numerous upsides. For instance, it guarantees that generated output will always adhere to the specified syntactic constraints, and reduces the need for ad-hoc parsing, retrying and prompting on top of LLMs. This facilitates the use of LLMs as part of larger pipelines or very specific tasks, without the need for fine-tuning or additional post-processing.

\begin{figure}[t]
    \def\scalef{1.0}

    {\fontsize{10pt}{0pt}\selectfont
    \textbf{Prompt:} A person encoded as JSON object:}

    \vspace{0.3em}

    {\textbf{Unconstrained Decoding:}\\[-0.3em]
    {
        \begin{spacing}{0.0}
            \fontsize{10pt}{0pt}\selectfont
            \dtoken{\{}{0}\dtoken{\ret}{0}\tcursor{$\downarrow$}{0}\\\dtoken{···}{0}\adtoken{·"}{0}\dtoken{name}{0}\dtoken{":}{0}\dtoken{·"}{0}\dtoken{John}{1}\dtoken{·Do}{2}\dtoken{e}{0}\dtoken{",}{0}\dtoken{\ret}{0}\\\dtoken{···}{0}\dtoken{·"}{0}\dtoken{age}{0}\dtoken{":}{0}\dtoken{·}{0}\dtoken{3}{0}\dtoken{2}{6}\dtoken{,}{0}\dtoken{\ret}{0}\\\dtoken{···}{0}\dtoken{·"}{0}\dtoken{gender}{6}\dtoken{":}{0}\dtoken{·"}{0}\dtoken{male}{0}\dtoken{",}{0}\dots

        \end{spacing}
    }
    }

    \vspace{1.0em}
    \vspace{0.8em}

    \textbf{Valid Tokens in Greedily Constrained JSON:}\\
    {
        \ttoken{[·\ret\tabchar]}\textit{(whitespace)} \hspace{0.2em} \ttoken{"}\textit{(quote)} \hspace{0.2em}
        \ttoken{\}}\textit{(closing brace)} \hspace{0.2em} 
    }\vspace{0.6em}

    {\textbf{Greedy Constraining induces sub-optimal tokenization:}\\[-0.5em]
    {
        \begin{spacing}{0.0}
            \fontsize{10pt}{0pt}\selectfont
            \dtoken{\{}{1}\dtoken{\ret}{1}\tcursor{$\downarrow$}{0}\\\dtoken{···}{1}\adtoken[my-blue]{\tabchar}{100}\dtoken{"}{1}\dtoken{name}{2}\dtoken{"}{100}\dtoken{\tabchar}{81}\dtoken{:}{4}\dtoken{\tabchar}{31}\dtoken{"}{1}\dtoken{John}{2}\dtoken{·Do}{8}\dtoken{e}{1}\dtoken{"}{100}\dtoken{\ret}{13}\\\dtoken{\tabchar}{6}\dtoken{,}{55}\dtoken{\tabchar}{10}\dtoken{"}{1}\dtoken{age}{1}\dtoken{"}{1}\dtoken{\tabchar}{1}\dtoken{:}{2}\dtoken{\tabchar}{1}\dtoken{3}{2}\dtoken{5}{8}\dtoken{\ret}{2}\\\dtoken{\tabchar}{1}\dtoken{,}{1}\dtoken{\tabchar}{1}\dtoken{"}{1}\dtoken{·occupation}{17}\dtoken{"}{2}\dtoken{\tabchar}{3}\dtoken{:}{1}\dtoken{\tabchar}{1}\dtoken{"}{1}\dtoken{So}{1}\dtoken{ftware}{1}\dots
        \end{spacing}
    }
    }
    \vspace{1.0em}
    \caption{Greedy (overly-invasive) constraining of LLMs can distort tokenization, leading to different output than with unconstrained decoding, even in the case where unconstrained generation would produce valid output for the same prompt.
        Gray boxes represent vocabulary tokens, orange hue is proportional to perplexity.}
    \label{fig:intro}
\end{figure}

\paragraph{Key Challenge: Token Misalignment} Since LLM sub-word token do not align directly with most given syntactic constraints, the key challenge in constrained decoding is to interface the LLM vocabulary with a syntactic constraint like \emph{all output should be valid JSON}.
We showcase this in \cref{fig:intro}: While in an unconstrained setting, the LLM picks \ttoken{·"} as the fourth token during generation, naively restricting the LLM to only immediatly valid JSON grammar terminals like just \ttoken{"} or \ttoken{·}, leads to the less optimal choice of \ttoken{\tabchar} instead. By introducing such sub-optimal tokens, the distribution of a badly-constrained LLM can easily diverge from the unconstrained case, leading to a significant decrease in reasoning performance and therefore downstream accuracy. Here, the naively constrained LLM produces various high perplexity tokens, indicating that the model likely would not have chosen them otherwise (highlighted in \cref{fig:intro}). This is because naive constraining does not account for \emph{bridge tokens} that may span multiple parser terminals in the underlying grammar (e.g., whitespace and double quotes, in this example). While existing work on code generation has made this observation before \cite{PoesiaP00SMG22}, solving this problem efficiently remains challenging, as the online computation of all bridge tokens at each decoding step, can be too costly in high-throughput environments.

\paragraph{This work: Efficient, Minimally-Invasive Constraining}
In this work, we study the token misalignment problem outlined above, and examine its consequences for constrained decoding, showing empirically that misalignment can lead to a significant decrease in downstream accuracy.
Based on this observation, we propose the notion of \emph{minimally invasive constrained decoding}: A form of constraining that enforces a grammar, but also intervenes as little as possible during generation, avoiding token misalignment and optimizing for faithful, low-perplexity model output.

Based on this, we propose a novel constrained decoding algorithm, \tool, which can enforce context-free grammars in a minimally-invasive way. In contrast to existing methods, \tool is highly efficient and incurs little to no overhead, and in many cases even increases the throughput of LLM inference over unconstrained generation, by leveraging pre-computation \citep{WillardL23} and a novel speculative decoding procedure for constrained decoding.
We compare \tool with other approaches in \cref{tab:constrained_decoding_methods}.

\textbf{Main contributions}
In summary, our key contributions are:
\begin{itemize}
    \item We identify the challenges of constrained decoding, most notably the correct and efficient alignment of sub-word tokens and grammar terminals (\cref{sec:token_misalignment}).
    \item We propose \tool, a novel constrained decoding algorithm, that addresses token misalignment and leverages pre-computation and speculative decoding for very low overhead generation (\cref{sec:technical}).
    \item An extensive evaluation thats shows that \tool is minimally-invasive, low-overhead, significantly outperforms other methods, and even exceeds unconstrained generation throughput in many cases (\cref{sec:eval}).
\end{itemize}

\section{Challenges of Constrained Decoding} \label{sec:token_misalignment}

\begin{table}[t]
\centering
\footnotesize
\caption{Overview of different constrained decoding methods
}
\begin{threeparttable}
\begin{tabular}{lccccc}
\toprule
                &Regex              & CFG               & \shortstack{Pre-\\Computed} & \shortstack{Minimally\\Invasive}\\
\midrule
\lmql           & \yesonline                    & \no               & \no         & \no\\
\guidance       & \yesonline                    & \partialonline    & \no         & \yespartial\tnote{$*$}\\
\outlines       & \yesoffline         & \yesonline        & \partialonline\tnote{$\dag$}      & \no\\
\picard         & \yes  & \yesonline        & \no                              & \no\\
\synchromesh    & \yes  & \yesonline        & \no                               & \yes\\
\llamacpp       & \yesonline                    & \yesonline        & \no & \yes\tnote{$+$}\\
\tgcd & \yesonline                    & \yesonline        & \no & \yes\\
\midrule
\tool (ours)           & \yes & \yes & \yes & \yes\\ 
\bottomrule
\end{tabular}
\begin{tablenotes}
    \footnotesize
    \item[$*$] Boundary token healing, \textsuperscript{$+$} Up to implementation. We observe violations in some cases, \textsuperscript{$\dag$} For regex.
\end{tablenotes}
\end{threeparttable}
\normalsize
\label{tab:constrained_decoding_methods}
\end{table}

We first introduce the required background and highlight the challenges of efficient and token-aligned constraining.

\paragraph{Large Language Models} (LLMs) are machine learning models trained to complete a given text prompt. The current generation of these models operate on sub-word tokens, such as the commonly used Byte-Pair Encoding (BPE) \citep{SennrichHB16a,KudoR18}. Conditioned on a sequence of input tokens $\ltok_1, \dots, \ltok_n$ the model computes a probability distribution over the next token,
from which the next token is decoded, i.e. chosen or sampled.
This process is repeated for each token.
In this context, we denote the set of all tokens, the vocabulary, as \voc.

\paragraph{Constrained Decoding}
\Cref{alg:constrained_decoding} shows the general outline used by most constrained decoding approaches.
A checker $C$, e.g., a parser or regex checker, is used to ensure that given an input $x$ the generated output $o$ adheres to the constraint. For each new token, $C$ is first updated with the latest generated sequence and then used to generate a mask $\vm$. This mask enforces that the next token must be a valid continuation.
Each time the mask in \cref{alg:constrained_decoding} rejects a token that would otherwise have been chosen, we say that the constrained decoding algorithm \emph{intervenes} in the decoding process.
We can therefore say that an algorithm that intervenes as little as possible is \emph{minimally invasive}:

\begin{algorithm}[t]
    \caption{Constrained Decoding} \label{alg:constrained_decoding}
    \begin{algorithmic}[1]
    \REQUIRE Checker $C$, LLM $f$, Tokenized Prompt $x$
    \ENSURE Completion $o$ adhering to $C$
    \STATE $o \leftarrow []$
    \STATE $C.\ttt{init}()$ \alglinelabel{alg:constrained_decoding:init}
    \LOOP
    \STATE $C.\ttt{update}(o)$ \alglinelabel{alg:constrained_decoding:update} \COMMENT{advance state of $C$}
    \STATE $\vm \leftarrow C.\ttt{mask}()$ \alglinelabel{alg:constrained_decoding:mask} \COMMENT{compute mask}
    \STATE $\vv \leftarrow f(x + o)$ \COMMENT{compute logits}
    \STATE $\vv' \leftarrow \vm \odot \vv'$     
    \STATE $t \leftarrow \ttt{decode}(\alpha')$ \COMMENT{e.g., argmax or sample}
    \STATE \textbf{if} $t = EOS$ \textbf{then break}
    \STATE $o.\ttt{append}(t)$
    \ENDLOOP
    \STATE \textbf{return} $o$ \COMMENT{optionally detokenize}
    \end{algorithmic}
\end{algorithm}

\begin{definition}[Minimally invasive]
We consider a constrained decoding method \emph{minimally invasive}, if every valid output that can be generated by an unconstrained model (for any given prompt), is also generated by the constrained model, given the same prompt.
\end{definition}

Replacing the naive constraining in \cref{fig:intro} with a minimally invasive method, would lead to the same output as with unconstrained generation, up to the final closing \ttoken{\}} bracket of the JSON object. After that, an unconstrained model continues to generate text, which will not part of the JSON object. At this point, a minimally invasive decoder, however, stops decoding by forcing an the \ttoken{\eos} token, to ensure that the output remains valid JSON.

As this example shows, non-minimally naive constrained decoding (as in \cref{fig:intro}) can lead to different overall output, compared to unconstrained generation. This in turn, can affect the model's task accuracy as we will show in \cref{sec:eval}, for a JSON-encoded version of the \textsc{GSM8K} dataset \citep{cobbe2021training}. There, naive constraining can reduce accuracy from $41.5\%$ in the unconstrained case to $30.8\%$, while minimally invasive constraining actually achieves a slight increase in accuracy to $41.8\%$ (five-shot, \emph{Mistral 7B} \citet{jiang2023mistral}).

Next, we discuss the broad categories of existing constrained decoding approaches. Current approaches implement one or multiple of the following paradigms:
i) \emph{Regex-Based}, ii) \emph{Online Parser-Guided} and iii) \emph{Template-Based} approaches.

\paragraph{Regex-Based} Decoding limited to regular expressions is supported by many frameworks (e.g., \lmql \citep{Beurer-Kellner023}, \outlines \citep{WillardL23}, \guidance \citep{guidance}, \llamacpp\citep{llamacpp}) and typically does not suffer from the token misalignment problem as it is simpler to check if a token is a legal continuation or not. For this, \citet{WillardL23} also proposed an algorithm to pre-compute a regex checker for the model vocabulary offline, to be more efficient during inference. %

\paragraph{Online Parser-Guided} refers to running a parser and scanner in lock-step with an LLM, and then computing online, which tokens are valid continuations in each step.
Such algorithms (\picard \citep{ScholakSB21}, \tgcd \citep{GengJP023}, \llamacpp) can support full CFGs and as \citet{PoesiaP00SMG22} (\synchromesh) demonstrated, can be built to support bridge tokens and thus to be minimally invasive. 
However, all of these approaches produce comparatively high inference overhead, since, in the worst case, they have to check the entire model vocabulary at each step.

\begin{figure}[t]
    \def\scalef{1.0}

    {\fontsize{10pt}{0pt}\selectfont
    \textbf{Prompt:} Tell me one sentence about Thomas Chapais.\ret\ret A: (Response In JSON)\vspace{0.5em}\\
    }
    {\textbf{(1a) Templated, Multi-Line} \hfill \footnotesize{\textit{Perplexity: $24.50$}}\\[-0.5em]
    {
        \begin{spacing}{0.0}
            \abtoken{\ret}{1}\abtoken{\{}{14}\abtoken{\ret}{1}\abtoken{···}{10}\abtoken{·"}{1}\abtoken{reason}{54}\abtoken{ing}{19}\abtoken{":}{3}\abtoken{·"}{3}\dtoken{Th}{16}\dtoken{omas}{0}\dtoken{·Chap}{1}\dtoken{ais}{0}\dtoken{·is}{8}\dtoken{·a}{4}\dtoken{·Canadian}{12}\dtoken{·politician}{17}\dots
        \end{spacing}
    }
    \vspace{2em}
    }
    {\textbf{(1b) Templated, Single-Line} \hfill \footnotesize{\textit{Perplexity: $26.75$}}\\[-0.5em]
    {
        \begin{spacing}{0.0}
            \abtoken{·\{}{21}\abtoken{·"}{11}\abtoken{reason}{53}\abtoken{ing}{22}\abtoken{":}{4}\abtoken{·"}{3}\dtoken{I}{16}\dtoken{·don}{9}\dtoken{'}{0}\dtoken{t}{0}\dtoken{·know}{2}\dtoken{·who}{15}\dtoken{·Thomas}{8}\dtoken{·Chap}{0}\dtoken{ais}{0}\dtoken{·is}{0}\dots
        \end{spacing}
    }
    \vspace{2em}
    }
    {
        {\textbf{(2) Naturalized Template Output} \hfill \footnotesize{\textit{Perplexity: $49.39$}}\\[-0.5em]
        \begin{spacing}{0.0}
            \dtoken{·\{}{18}\dtoken{·"}{10}\dtoken{re}{40}\dtoken{as}{57}\dtoken{o}{26}\dtoken{ning}{66}\dtoken{":}{3}\dtoken{·"}{3}
            \dtoken{I}{16}\dtoken{·don}{9}\dtoken{'}{0}\dtoken{t}{0}\dtoken{·know}{2}\dtoken{·who}{15}\dtoken{·Thomas}{8}\dtoken{·Chap}{0}\dtoken{ais}{0}\dtoken{·is}{0}\dots
        \end{spacing}
    }
    \vspace{2em}
    }
    {\textbf{(3) Output with Naturalized Template} \hfill \footnotesize{\textit{Perplexity: $4.17$}}\\[-0.5em]
    {
        \begin{spacing}{0.0}
            \fontsize{10pt}{0pt}\selectfont
            \dtoken{·\{}{21}\dtoken{·"}{11}\dtoken{re}{46}\dtoken{as}{65}\dtoken{o}{30}\dtoken{ning}{75}\dtoken{":}{3}$\rightsquigarrow$\dtoken{·"}{2}\dtoken{I}{16}\dtoken{·don}{8}\dtoken{'}{0}\dtoken{t}{0}\dtoken{·know}{2}\dtoken{·him}{14}\dtoken{"}{8}\dtoken{·\}}{1}
        \end{spacing}
    }
    }
    \vspace{1.0em}
    \caption{Template-based tokens, marked as \abtoken{ }{21}, force unnatural tokenization and formatting, which can lead to different outputs and increased perplexity. Gray boxes represent vocabulary tokens, hue is proportional to perplexity.}
    \label{fig:misaligned}
\end{figure}

\paragraph{Template-Based Approaches} 
Since constrained generation can add additional overhead during inference, \guidance and \lmql, propose a template-based approach, where some structure is fixed, and only parts of the output are sampled under a regex constraints.
For instance, templated generation can be used to implement schema-driven JSON, where fields are fixed.
Template-based decoding is efficient as templated tokens can be added deterministically during generation, without invoking the LLM, thereby requiring less model forward passes.
However, this form of acceleration also has its downside, as discussed next.

\paragraph{Template-Induced Misalignment}
To insert templated tokens without invoking the model, an external tokenizer has to be used to translate template text into tokens. We showcase this in \cref{fig:misaligned}, where we use the template-based \guidance with \emph{Mistral-7B}, to generate text in JSON format, and compare it to unconstrained generation. (1a) and (1b) already show that depending on the concrete phrasing of a template (e.g., whitespace, formatting), output can vary significantly. To compare to unconstrained generation, we use the model to also generate the template text, but without imposing a fixed tokenization (algorithm in \cref{app:retokenization}). We then generate output with this \emph{naturalized} template (3) to contrast it with template-induced tokenization.

Overall, template-based outputs exhibit clearly different outputs and much higher perplexities (perplexities $24.50-26.75$), when compared to unconstrained generation (perplexity $4.17$). Further, we observe that naturalizing the full template-based output under the model-preferred tokenization (2) in \cref{fig:misaligned}, results in a form of perplexity explosion ($49.39$), indicating that without invasive constraining, the model is highly unlikely to generate such output. While perplexity and output differences are not directly indicative of output quality, our evaluation in \cref{sec:task_accuracy_discussion} extends on this experiment, and indeed shows that invasive template-based generation can lead to a significant drop in task accuracy, compared to unconstrained and less invasive approaches.

\paragraph{Token Healing} To reduce the impact of template-induced misalignment, \guidance implements \emph{token healing} \citep{tokenhealing}, a method that attempts to improve tokenization at the transition points between templated and non-templated tokens. Token healing truncates the prompt to the second-to-last token boundary, and enforces the rest of the prompt as a prefix on generation.
This can be effective at avoiding some tokenization issues and integrating bridge tokens, yet, most templated tokens remain fixed as demonstrated in \cref{fig:misaligned}, where token healing is already in effect, and still leads to significant differences in output.

\paragraph{Key Challenge}
This section, as summarized in \cref{tab:constrained_decoding_methods}, demonstrates that the key challenge in constrained decoding is to find a method that allows i) expressivity for regex, context-free grammars and templated decoding, ii) is minimally invasive in all settings and iii) has low inference overhead. Next, we discuss \tool, a novel constrained decoding algorithm, that fits these requirements.

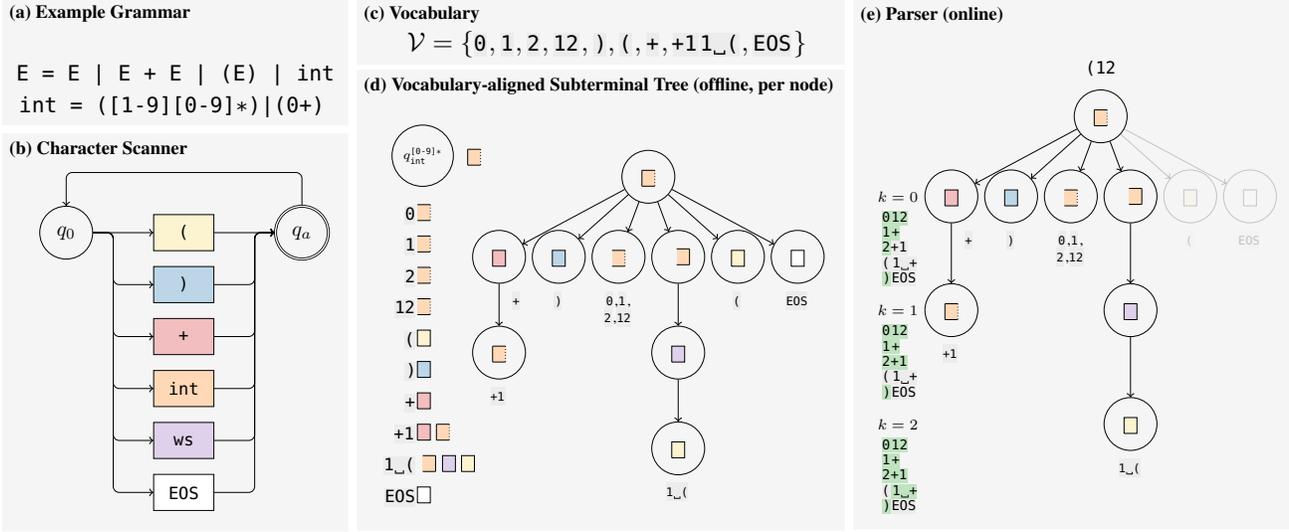
\begin{figure*}
\centering 
\begin{minipage} {0.27\textwidth} 
\centering
\bgbox{(a) Example Grammar}{
\ttt{E = E | E + E | (E) | int}
\ttt{int = ([1-9][0-9]*)|(0+)}
}

\bgbox{(b) Character Scanner}{
\scalebox{0.8}{
\begin{tikzpicture}[
rnode/.style={%
rectangle,
minimum width = 10mm,
minimum height = 6mm,
draw=black,
fill=white
},
]
\def\sep{2.5mm}
\node at (2, 1.1) {};
\node[state] (start) at (0, 0) {$q_0$};
\node[rnode, right=10mm of start, fill=my-yellow] (term0) {\texttt{(}};
\node[state, accepting, right=10mm of term0] (end) {$q_a$};
\node[rnode, below=\sep of term0, fill=my-blue] (term1) {\texttt{)}};
\node[rnode, below=\sep of term1, fill=my-red] (term2) {\texttt{+}};
\node[rnode, below=\sep of term2, fill=my-orange, fill=my-orange] (term3) {\texttt{int}};
\node[rnode, below=\sep of term3, fill=my-purple] (term4) {\texttt{ws}};
\node[rnode, below=\sep of term4, fill=white] (term5) {\texttt{\eos}};

\draw[->] (start) edge node[above] {} (term0);
\draw[->] (term0) edge node[above] {} (end);

\coordinate (split_in) at ($(start)!0.4!(term0)$);
\draw[->, rounded corners=1mm] (start) -- (split_in) -- (split_in |- term1)  -- (term1);
\draw[->, rounded corners=1mm] (start) -- (split_in) -- (split_in |- term2)  -- (term2);
\draw[->, rounded corners=1mm] (start) -- (split_in) -- (split_in |- term3)  -- (term3);
\draw[->, rounded corners=1mm] (start) -- (split_in) -- (split_in |- term4)  -- (term4);
\draw[->, rounded corners=1mm] (start) -- (split_in) -- (split_in |- term5)  -- (term5);

\coordinate (split_out) at ($(term0)!0.6!(end)$);
\draw[->, rounded corners=1mm] (term1) -- (split_out |- term1) --  (split_out)  -- (end);
\draw[->, rounded corners=1mm] (term2) -- (split_out |- term2) --  (split_out)  -- (end);
\draw[->, rounded corners=1mm] (term3) -- (split_out |- term3) --  (split_out)  -- (end);
\draw[->, rounded corners=1mm] (term4) -- (split_out |- term4) --  (split_out)  -- (end);
\draw[->, rounded corners=1mm] (term5) -- (split_out |- term5) --  (split_out)  -- (end);

\draw[->, rounded corners=1mm] (end) -- ($ (end)+(0,1) $) --  ($ (start) + (0,1) $)  -- (start);

\node[below=-0.3mm of term5] {};

\end{tikzpicture}
}}

\end{minipage}
\hfill
\begin{minipage}{0.38\textwidth}
\centering
\bgbox{(c) Vocabulary}
{$\voc = \{\ttoken{0}, \ttoken{1}, \ttoken{2}, \ttoken{12}, \ttoken{)}, \ttoken{(}, \ttoken{+}, \ttoken{+1} \ttoken{1\verbvisiblespace{}(}, \ttoken{\eos} \}$}
\bgbox{(d) Vocabulary-aligned Subterminal Tree (offline, per node)}
{
\scalebox{0.8}{
\begin{tikzpicture}
    \node[state] (s1) at (0, 0) {\tiny $q^{\ttt{[0-9]*}}_{\ttt{int}}$};
    \node[right=1mm of s1] (s1domino) {\terminalstartme{my-orange}};
\node[below left=2.75mm and 4mm of s1, anchor=north west] (s1dominos) {
\begin{tabular}{@{}r@{\vspace{1mm}}l@{}}
\ttoken{0} & \terminalmiddleme{my-orange}\\
\ttoken{1} & \terminalmiddleme{my-orange}\\
\ttoken{2} & \terminalmiddleme{my-orange}\\
\ttoken{12} & \terminalmiddleme{my-orange}\\
\ttoken{(} & \terminal{my-yellow} \\
\ttoken{)} & \terminal{my-blue} \\
\ttoken{+} & \terminal{my-red} \\
\ttoken{+1} & \terminal{my-red} \terminalstartme{my-orange}\\
\ttoken{1\verbvisiblespace{}(} & \terminalend{my-orange} \terminal{my-purple} \terminal{my-yellow} \\
\ttoken{\eos} & \terminal{white}  
\end{tabular}
    };

\node[state] (t0) at (3.75, -0.35) {\terminalstartme{my-orange}};

\node[state, below left  = 7mm and 18.5mm of t0] (t24) {\terminal{my-red}};
\node[state, below right = 7mm and 18.5mm of t0] (t25) {\terminal{white}};

\node[state, left = 1mm of t25] (t22) {\terminal{my-yellow}};
\node[state, right = 1mm of t24] (t23) {\terminal{my-blue}};

\node[state, left  = 1mm of t22] (t20) {\terminalend{my-orange}};
\node[state, right = 1mm of t23] (t21) {\terminalmiddleme{my-orange}};

\node[state, below = 7mm of t24] (t30) {\terminalstartme{my-orange}};
\node[state, below = 7mm of t20] (t31) {\terminal{my-purple}};
\node[state, below = 7mm of t31] (t40) {\terminal{my-yellow}};

\draw[->] (t0) edge node[below] {} (t20);
\draw[->] (t0) edge node[below] {} (t21);
\draw[->] (t0) edge node[below] {} (t22);
\draw[->] (t0) edge node[below] {} (t23);
\draw[->] (t0) edge node[below] {} (t24);
\draw[->] (t0) edge node[below] {} (t25);

\draw[->] (t24) edge node[below] {} (t30);
\draw[->] (t20) edge node[below] {} (t31);
\draw[->] (t31) edge node[below] {} (t40);

\node[below=0mm of t22] {\scriptsize\ttoken{(}};
\node[below=0mm of t23] (lt23) {\scriptsize\ttoken{)}};
\node[left =3mm of lt23] {\scriptsize\ttoken{+}};
\node[below=0mm of t25] {\scriptsize\ttoken{\eos}};
\node[below=0mm of t21] {\scriptsize\shortstack{\ttoken{0},\ttoken{1},\\[-0.5em]\ttoken{2},\ttoken{12}}};

\node[below=0mm of t40] {\scriptsize\ttoken{1\verbvisiblespace{}(}};
\node[below=0mm of t30] {\scriptsize\ttoken{+1}};

\end{tikzpicture}
}}
\end{minipage}
\hfill
\begin{minipage}{0.34\textwidth}
\bgbox{(e) Parser (online)}
{
\scalebox{0.8}{
\begin{tikzpicture}
    
\node[state] (t0) at (3.75, -0.35) {\terminalstartme{my-orange}};

\node[state, below left  = 7mm and 18.5mm of t0] (t24) {\terminal{my-red}};
\node[state, opacity=0.2,  below right = 7mm and 18.5mm of t0] (t25) {\terminal{white}};

\node[state, opacity=0.2, left = 1mm of t25] (t22) {\terminal{my-yellow}};
\node[state, right = 1mm of t24] (t23) {\terminal{my-blue}};

\node[state, left  = 1mm of t22] (t20) {\terminalend{my-orange}};
\node[state, right = 1mm of t23] (t21) {\terminalmiddleme{my-orange}};

\node[state, below = 10mm of t24] (t30) {\terminalstartme{my-orange}};
\node[state, below = 10mm of t20] (t31) {\terminal{my-purple}};
\node[state, below = 10mm of t31] (t40) {\terminal{my-yellow}};

\draw[->] (t0) edge node[below] {} (t20);
\draw[->] (t0) edge node[below] {} (t21);
\draw[->, opacity=0.2 ] (t0) edge node[below] {} (t22);
\draw[->] (t0) edge node[below] {} (t23);
\draw[->] (t0) edge node[below] {} (t24);
\draw[->, opacity=0.2 ] (t0) edge node[below] {} (t25);

\draw[->] (t24) edge node[below] {} (t30);
\draw[->] (t20) edge node[below] {} (t31);
\draw[->] (t31) edge node[below] {} (t40);

\node[below=0mm of t22, opacity=0.2 ] {\scriptsize\ttoken{(}};
\node[below=0mm of t23] (lt23) {\scriptsize\ttoken{)}};
\node[left =3mm of lt23] {\scriptsize\ttoken{+}};
\node[below=0mm of t25, opacity=0.2 ] {\scriptsize\ttoken{\eos}};
\node[below=0mm of t21] {\scriptsize\shortstack{\ttoken{0},\ttoken{1},\\[-0.5em]\ttoken{2},\ttoken{12}}};

\node[below=0mm of t40] {\scriptsize\ttoken{1\verbvisiblespace{}(}};
\node[below=0mm of t30] {\scriptsize\ttoken{+1}};

\node[above =1mm of t0] {\ttt{(12}};
\node[left=28mm of t0] (x) {};
\node (k0) at (x|-t20) {\scriptsize$k=0$};

\node[below=-1mm of k0] (lk0) {
\scalebox{0.8}{
\renewcommand{\arraystretch}{0.7}
\begin{tabular}{@{}l@{}l@{}}
\ttoken[accept]{0} & \ttoken[accept]{12}\\
\ttoken[accept]{1} & \ttoken[accept]{+}\\
\ttoken[accept]{2} & \ttoken{+1}\\
\ttoken{(} & \ttoken{1\verbvisiblespace{}+}\\
\ttoken[accept]{)} & \ttoken{\eos}\\
\end{tabular}
}
};

\node (k1) at (x|-t30) {\scriptsize$k=1$};

\node[below=-1mm of k1] (lk1) {
\scalebox{0.8}{
\renewcommand{\arraystretch}{0.7}
\begin{tabular}{@{}l@{}l@{}}
\ttoken[accept]{0} & \ttoken[accept]{12}\\
\ttoken[accept]{1} & \ttoken[accept]{+}\\
\ttoken[accept]{2} & \ttoken[accept]{+1}\\
\ttoken{(} & \ttoken{1\verbvisiblespace{}+}\\
\ttoken[accept]{)} & \ttoken{\eos}\\
\end{tabular}
}
};

\node (k2) at (x|-t40) {\scriptsize$k=2$};
\node[below=-1mm of k2] (lk2) {
\scalebox{0.8}{
\renewcommand{\arraystretch}{0.7}
\begin{tabular}{@{}l@{}l@{}}
\ttoken[accept]{0} & \ttoken[accept]{12}\\
\ttoken[accept]{1} & \ttoken[accept]{+}\\
\ttoken[accept]{2} & \ttoken[accept]{+1}\\
\ttoken{(} & \ttoken[accept]{1\verbvisiblespace{}+}\\
\ttoken[accept]{)} & \ttoken{\eos}\\
\end{tabular}
}
};

\end{tikzpicture}
}}
\end{minipage}
\vspace{-1em}
\caption{Running example and overview of \tool. (a) shows an example grammar, (b) the character level NFA for this language, (d) one of the per-state subterminal trees for the grammar in (c). (e) shows how a parser can be used to prune this tree at inference time and obtain token masks efficiently by traversing the tree.} \label{fig:overview}
\vspace{-2mm}
\end{figure*}

\section{Efficient Aligned Constrained Decoding} \label{sec:technical}

In this section, we address the above challenges by introducing \tool, a fast minimally-invasive constrained decoding algorithm, showcased in \cref{fig:overview}. First, \cref{sec:technical:preliminaries} discusses the necessary preliminaries. Then, in \cref{sec:technical:scanner}-\cref{sec:technical:summary} we present the main algorithm and then discuss further details.
\subsection{Preliminaries} \label{sec:technical:preliminaries}

\paragraph{Formal Languages} 
A formal language is a set $L$ of finite strings over a given alphabet $\Sigma$.
A language is called regular if it can be described by a regular expression.
A language is context-free if it can be described by a context-free grammar (CFG). %
Such a grammar is described by a set of production rules $A = \alpha B \gamma$, where upper case names ($A$, $B$) are non-terminals that are recursively extended and lower case greek characters (e.g., $\alpha$, $\beta$) are terminals that are part of the language. For an example, see the grammar in \cref{fig:overview}~(a), with the non-terminal \ttt{E} and terminals \ttt{int}, \ttt{(}, \ttt{)}, \ttt{+}, which are defined either by a regex or a literal string.

\begin{figure}[t]
    \centering
    \begin{tikzpicture}[
state/.style={%
    circle,
    minimum size = 8mm,
    draw=black,
},
]
    \node[state] (q0) at (0, 0) {\tiny \texttt{int}};
    \node[state, accepting, left=8mm  of q0] (q1) {\tiny \texttt{0*}};
    \node[state, accepting, right=8mm  of q0] (q2) {\tiny \texttt{[0-9]*}};
    \draw[->] (q0) edge node[above] {\tiny \texttt{0}} (q1);
    \draw[->] (q0) edge node[above] {\tiny \texttt{[1-9]}} (q2);
    \draw[->] (q1) edge [loop left] node {\tiny \texttt{0}} (q1);
    \draw[->] (q2) edge [loop right] node {\tiny \texttt{[0-9]*}} (q2);

\end{tikzpicture}
\vspace{-1em}
\caption{NFA for the \ttt{int} terminal from \cref{fig:overview}~(a). Traversed from node \ttt{int} this NFA accepts all legal inputs for the terminal.} \label{fig:nfa}
\vspace{-2em}
\end{figure}
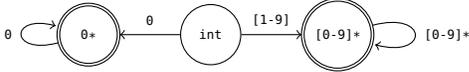

All regular languages can be recognized by a non-deterministic finite automata (NFA).
A NFA is a set of states with a start state and a set of accepting states.
Connected with transitions that are labeled with a character or the empty string $\varepsilon$.
We traverse or execute an NFA by starting in the start state and whenever we read a character follow the appropriate transitions or any $\varepsilon$-transitions.
For a regular expression, we can construct a NFA \citep{McNaughtonY60,Thompson68}, that when fed a string, character by character, ends in a set of state including at least one accepting state, if and only if the string matches the regular expression.
To illustrate this approach consider the grammar given in \cref{fig:overview}~(a). The terminal $\ttt{int}$ is given by a regex that allows any positive integer (not starting with leading zeros) or one or more zeros. \cref{fig:nfa} shows an NFA for this regex.

\subsection{Character Scanner} \label{sec:technical:scanner}
Like classical parsers, \tool separates the CFG recognition into a parser and a scanner (or lexer).
The parser enforces the high-level structure of the language, e.g., the rules in a context-free grammar, and the scanner enforces the low-level structure, i.e., the regular expressions of the terminals.
The key idea is that any legal program in a context-free grammar is a sequence of terminals, or formally:

\begin{lemma}
Let $L_{G}$ be the language described by a CFG $G$.
Further, let $r_1, \ldots, r_n$ be the regular expressions of the terminals of $G$ and the $r_{\eos} = \$$. Then, it holds that:
\begin{itemize}[noitemsep]
\item  The union of these regular expressions $r = r_1|\ldots|r_n|r_{\eos}$ matches any terminal in $G$.
\item  The regular expression $R = r+$ matches all non-empty sequences of terminals in the language.
\item The language $L_R$ described by $R$ contains all legal programs in $G$, i.e., $L_G \subseteq L_R$, but also
\item $L_R$ contains some illegal programs, i.e., $L_R \not\subseteq L_{G}$.
\end{itemize}
\end{lemma}

To construct an NFA for $L_R$, we construct the NFA for the regex $r_i$ for each terminal in the grammar.
From the accepting states of the invidual terminal NFAs, we add an $\varepsilon$-transition to a single NFA accepting state $q_a$.
Similarly, we add a transition from the start state $q_0$ to the start state of each DFA.
Finally, we add a transition from $q_0$ to $q_a$, to allow for the chaining of multiple tokens.
This is the standard disjuction construction for regex NFAs, however, we do this explictly to track the sub-automata correspoding to each terminal.
This construction is showcased in \cref{fig:overview}~(b) for the grammar in \cref{fig:overview}~(a). There, the boxes in the middle correspond to individual NFAs like \ttt{int} as shown in \cref{fig:nfa}.

\subsection{Vocabulary-Aligned Subterminal Tree}
Using a scanner $S$ at generation time can enforce that the generated string will be in $L_R$, but not neccesarily in $L_{G}$. To ensure this, we need to also run a parser $P$ on the output as it is generated and dynamically allow and disallow some of the transitions in $S$, according to the current parser state.

For this, we first lift $S$ from the character level to the (sub)terminal level used by the parser. To do so, for each node in $S$ we follow the transitions for each token in the vocabulary and enumerate all reachable states.
Particularly, we track which terminal NFAs are partially or fully traversed.
As terminals are not neccesarily aligned with the token vocabulary, we introduce the notion of a \emph{subterminal} as a part of a terminal NFA. In particular, for terminal $\alpha$, e.g., \ttt{int}, we say that $S$ reads a:
\begin{itemize}[nosep]
    \item \textbf{Full terminal} if it passes $q_0$ and reaches an accepting state $q_\alpha$ in the NFA for the terminal. We denote this as \terminal{my-lightgray}. If we end in an accepting state, that also allows further transitions, such as both accepting states in \cref{fig:nfa}, we also consider this a full terminal, but allow further subterminals within the terminal NFA (\terminalstartme{my-lightgray}).
    \item \textbf{Start subterminal} if starting from $q_0$ we reach a non-accepting but valid state $q_\alpha$ for the NFA of the terminal. We denote this as \terminalstart{my-lightgray}.
    \item \textbf{End subterminal} if starting from a state $q_\alpha$ within the NFA for $\alpha$ we reach an accepting state for $\alpha$. We denote this as \terminalend{my-lightgray}.
    \item \textbf{Continuation subterminal} if starting from a state $q_\alpha$ within the NFA for $\alpha$ we reach another non-accepting state for $\alpha$. We denote this as \terminalmiddle{my-lightgray}.
\end{itemize}

We visit every state $q$ in $S$, obtain the current (sub)terminal $\alpha$ and, for each vocabulary token $\ltok \in \voc$, enumerate all possible subterminal sequences $\{\alpha_1^j, \dots, \alpha^j_{m_j}\}_j$. Typically there is only one such sequence unless there is ambiguity in the grammar, e.g., in $C$-style languages we can not be sure if we are reading a variable name of a keyword. Ignoring these edge cases, we show an example of this on the left hand side of \cref{fig:overview}~(d) for the vocabulary given in \cref{fig:overview}~(c), where we visualize the subterminals in the previously introduced box notation with colors correspoding to the terminals in \cref{fig:overview}~(b). Note that there may be tokens in the vocabulary for which no continuation is possible, e.g., if we add an \ttoken{a} token to the language.
After enumerating all these subterminal sequences, we organize them into a prefix tree $T_q$, where we attach the corresponding vocabulary tokens $\ltok$ as values to the nodes. We formalize this procedure in \cref{alg:terminal_tree}.

\begin{algorithm}[t]
    \caption{Construct Terminal Tree} \label{alg:terminal_tree}
    \begin{algorithmic}[1]
    \REQUIRE CFG $G$, Alphabet $\Sigma$, Vocabulary $\voc$
    \ENSURE Scanner $S$
    \STATE $\mathcal{T} = \{\}$
    \FORALL{$q \in S.\ttt{states}()$}
    \STATE $\alpha \leftarrow q.\ttt{subterminal}()$ \COMMENT{get current (sub)terminal}
    \FORALL{$\ltok \in \voc$}
    \STATE $\{\alpha_1^j, \dots, \alpha^j_{m_j}\}_j \leftarrow q.\ttt{traverse}(\ltok)$
    \STATE $\mathcal{T} \leftarrow \mathcal{T} \cup \{ (\alpha_1^j, \dots, \alpha^j_{m_j}), \ltok \}_j$
    \ENDFOR
    \STATE $T_q \leftarrow \ttt{PrefixTree}(\mathcal{T})$
    \ENDFOR
    \end{algorithmic}
\end{algorithm}

\subsection{Parser} \label{sec:technical:parser}
While subterminal trees can be pre-computed, we still need a parser at inference time, to disallow illegal continuations.
For example, consider the prefix tree \cref{fig:overview}~(d). If so far, we have observed the sequence \ttt{(12} and correctly advanced the parser and scanner to this state, we are in an \ttt{int} terminal \terminalmiddleme{my-orange}, that can still be extended further. Following the prefix tree in \cref{fig:overview}~(d) would permit continuations such as \ttoken{(}, which are clearly illegal in the grammar. Thus, we need to disallow these continuations dynamically by consulting the parser.

At inference time traversing the so-far generated sequence $o$ through the scanner and parser will result in a scanner state $S$ and parser state $P$.
The active state of $S$ will be a set of states $q_1, \dots, s_m$.
The active state of $P$ will be a parser that tracks rules that can match the output $o$ so far. 
For each of these nodes $q_i$ we retrive the corresponding subterminal tree $T_{q_i}$ and use the parser to check which of the possible continuations are legal. The depths to which we follow these terminal sequences in the prefix-tree is determined by the so-called \emph{lookahead parameter} $k$.

We showcase this in \cref{fig:overview}~(e), where so far we have read the input \ttt{(12}. The parser thus knows that it has seen the partial rule \ttt{(E)}, in which recusively \ttt{E} was initialized with the \ttt{int} terminal.
The scanner $S$ has been advanced similarly and has a single active state $s$ that corresponds to that for an \ttt{int} terminal \terminalmiddleme{my-orange}, as shown in \cref{fig:overview}~(d). In the corresponding prefix tree $T_s$ we can now check each outbound edge with the parser and find that the \ttoken{\eos} and \ttoken{(} can not produce legal (sub)terminal sequences, but all other tokens do. After traversing one level of this tree ($k=0$) this includes all number terminals, \ttoken{+} and \ttoken{)}. By increasing $k$ and checking further paths in the tree we can also include the tokens \ttoken{+1} and \ttoken{1\verbvisiblespace{}(} at $k=1$ and $k=2$ respectively.

\subsection{\tool} \label{sec:technical:summary}

Based on \cref{alg:constrained_decoding}, \tool implements constraining that leverages subterminal trees to efficiently check for legal continuations in \voc, that, by construction, line up with the current state, just like the pieces in a game of domino. 

For this, we compute the character scanner $S$ and corresponding prefix trees $T_q$ for all $q \in S$ before inference starts (offline). When \ttt{update} is called in \cref{alg:constrained_decoding}, we can then advance the scanner and parser state. 
When \ttt{mask} is invoked 
, we traverse the corresponding $T_q$ to the desired depth, and take the union over the associated tokens to compute the current mask. %
Here, $k=0$ is already sufficient to ensure that the generated output is in $L_G$. If we always traverse the full prefix-trees ($k=\infty$ or sufficently large), this approach is minimally invasive, as all valid tokens will eventually be reached and included in the mask.

Overall, this enables us to handle expressive constraining with far less overhead than fully online approaches, as the size of subterminal trees is much smaller than the size of the model vocabulary, which we would need to traverse otherwise. Similarly \tool can also be extended to other forms of constraining, e.g. to execute \guidance programs.

\paragraph{Further Optimizations}
On top of this, \tool also supports an optimization already present in \llamacpp \citep{llamacpp}'s fully online approach, which we term \emph{opportunistic masking}: Rather than computing the mask for the full logit vectors as in \cref{alg:constrained_decoding}, we can  first run the decode step of the LLM and then use the parser to check the model-proposed token first. Only if incorrect, we need to compute the rest of the mask, and thus let the LLM guide decoding.

To realize this in \tool, rather than traversing the trees $T_q$ from the root, we first determine the nodes linked to the proposed token, and only then check if there exists a parser-allowed path from the root to this node.
This is can be very effective when an LLM already naturally adheres to a grammar, and parser transitions are expensive. In \tool, oppurtunistic masking is enabled using a runtime flag.

We finally want to note, that \emph{token healing} can be implmented in a similar way to \guidance \citep{guidance}, i.e., by stripping the input back to the last token boundary and changing the beginning of the grammar to force a prefix. However, this means that the grammar needs to be recompiled for the current problem and can not neccesarily be shared between multiple instance. Note however, that this is of lesser concern in \tool, as it is only relevent for the first boundary with the prompt, where all other boundaries are embedded seamlessly into the grammar.

\subsection{Speculative Decoding} \label{sec:speculative}
Speculative Sampling \citep{ChenBILSJ23} is a technique to speed up the LLM sampling process, by using a smaller LLM to propose multiple tokens and then only evaluate the full large LLM to confirm this token choice. This is efficient, as the parallel nature of Transformer-based models \citet{vaswani2017attention}, allows to validate multiple tokens with a single forward pass, where rejected tokens can simply be discarded without the need to backtrack.
In \tool, we adopt a similar approach to further speed up inference, based on parser and scanner state.
At any time, the active scanner state is (largely) given by the most recently read subterminal $\alpha$ (or $\{\alpha_j\}_j$ if the NFA could be in multiple possible subterminals).
Similarly, we let $\beta$ denote some sub-state of the currently used parser, e.g., the currently applied rule. 
Conditioned on $\alpha,\beta$ we can then learn a simple, count-based model for speculative next token prediction:
\begin{equation*}
    \mathbb{P}(\ltok \mid \alpha, \beta) = \frac{\#\{\text{LLM chose $\ltok$ in state $(\alpha,\beta)$}\}}{\#\{ \text{reached state } (\alpha, \beta) \}}.
\end{equation*}

As structured languages often are very predictable and $\alpha,\beta$ can be strongly indicative of the next token, this mechanism can lead to massive speed-up during inference.
Further, as we learn these counts over the parser state $\beta$, we only learn to predict tokens that are legal in the language.

In practice, we parameterize $s$ tokens to be predicted this way at a time, if the $\mathbb{P}(\ltok \mid \alpha, \beta)$ is sufficiently large. This form of speculative decoding is independent of standard speculative decoding applied to the underlying LLM \citep{ChenBILSJ23}, and could even be applied jointly with it.

\begin{table*}
    \centering

    \caption{Task Accuracy of different constrained decoding methods on \emph{GSM8K} \citet{cobbe2021training} and \textsc{CoNLL2003} \citet{sang2003introduction} datasets (400 test samples). All experiments rely on 5-shot prompting with demonstrations taken from the training split.}

    \scalebox{1.0}{
    \begin{threeparttable}
        \footnotesize
    \renewcommand{\arraystretch}{1.2}
    \centering
    \begin{tabular}{llrllll}
    \toprule
    \textbf{Dataset} & \textbf{Model} & \textbf{Method} & \textbf{Accuracy} & \textbf{Well-Formed} & \textbf{Perplexity} & \textbf{Performance Impact} \\
    \midrule
    \multirow{10}{*}{GSM8K} &
    \multirow{5}{*}{\emph{Mistral 7B}} 
             & Unconstrained &                                   0.415 & 0.952 & 1.636 & 1.0$\times$\\ 
            && \guidance \citet{guidance} &                      0.345 & 0.960 & 1.624 & 0.98$\times$\\
            && \guidance\textsuperscript{WS}  \citet{guidance} & 0.403 & 0.976 & 1.737 & 0.54$\times$ \\
            && \lstinline|llama.cpp| \citet{llamacpp} &          0.375 & 0.973 & 1.751 & 0.80$\times$\\
            && \tool ($k=\infty$) &                      \textbf{0.418}& 0.968 & 1.739 & \textbf{1.77$\times$} \\ \cline{3-7}
    & \multirow{5}{*}{\emph{Llama-2 13B}} 
            & Unconstrained & \textbf{0.262} & 0.904 & 1.650 & 1.0$\times$ \\
            && \guidance  \citet{guidance} & 0.152 & 0.947 & 1.659 & 1.12$\times$\\
            && \guidance\textsuperscript{WS}  \citet{guidance} & 0.259 & 0.977 & 1.760 & 0.73$\times$\\
            && \lstinline|llama.cpp|  \citet{llamacpp} & 0.237 & 0.978 & 1.780 & 0.86$\times$\\
            && \tool ($k=\infty$) & \textbf{0.262} & 0.920 & 1.750 & \textbf{1.66$\times$} \\
    \midrule
    \multirow{10}{*}{CoNLL2003} &
    \multirow{5}{*}{\emph{Mistral 7B}} 
             & Unconstrained & \textbf{0.21} & 0.988 & 1.573 & 1.0$\times$\\
            && \guidance \citet{guidance} & 0.098 & 0.998 & 1.780 & 2.02$\times$\\
            && \guidance\textsuperscript{WS}  \citet{guidance} & 0.19 & 0.998 & 1.896 & 0.82$\times$\\
            && \lstinline|llama.cpp| \citet{llamacpp} & 0.117 & 0.995 & 1.560 & 0.80$\times$\\
            && \tool ($k=\infty$) & \textbf{0.21} & 0.988 & 1.902 & \textbf{2.66$\times$}\\ \cline{3-7}
    & \multirow{5}{*}{\emph{Llama-2 13B}} 
            & Unconstrained & \textbf{0.09} & 0.897 & 1.579 & 1.0$\times$\\
            && \guidance  \citet{guidance} & 0.062 & 1.000 & 1.820 & 2.18$\times$\\
            && \guidance\textsuperscript{WS} \citet{guidance} & 0.087 & 0.980 & 1.767 & 0.90$\times$\\
            && \lstinline|llama.cpp|  \citet{llamacpp} & 0.080 & 0.922 & 1.786 & 0.86$\times$\\
            && \tool ($k=\infty$) & \textbf{0.09} & 0.897 & 1.812 & \textbf{2.71$\times$}\\
    \bottomrule
    \end{tabular}
    \begin{tablenotes}
        \footnotesize
        \item[WS] \guidance CFG program with flexible whitespace and formatting.
    \end{tablenotes}
    \label{tab:task_accuracy}
    \end{threeparttable}
    }
\end{table*}

\newpage
\section{Experimental Evaluation} \label{sec:eval}

We evaluate \tool in terms of downstream task accuracy, 
compare its performance to multiple baselines and ablate key parameters such as $k$.

\paragraph{Setup} We evaluate on the \emph{Mistral 7B} \cite{jiang2023mistral} and the \emph{Llama-2 13B} \cite{touvron2023llama} language models. As inference backends, we rely on both, \lstinline|transformers| \cite{wolf2019huggingface} and \lstinline|llama.cpp| \cite{llamacpp} on  NVIDIA A100 40GB or H100 80GB GPUs. Because of its nature, we explicitly evaluate \tool in an offline setting, where all grammars are known ahead of time and do not vary across inference requests. 

\paragraph{Datasets} We assess downstream accuracy of different constraining methods with the \emph{GSM8K} \cite{cobbe2021training} benchmark for math reasoning and \emph{CoNLL-2003} \cite{sang2003introduction} for named-entity recognition (subset of 400 test samples). To examine the performance properties of different decoding methods, we compare their overhead over unconstrained decoding for different tasks, including the constrained generation of \emph{JSON}, \emph{JSON with Schema}, the \emph{C Programming Language}, \emph{XML with Schema} and more static, regex-based generation templates similar to \guidance or \lmql programs with simple structure.

\paragraph{Baselines} We consider the following baselines: 

\begin{itemize}[nosep]
    \item \textbf{Unconstrained Generation} We generate output without any form of constraints, using the same prompts and inference backend.
    \item \textbf{\guidance Programs} \citep{guidance} We construct \guidance programs to generate output in the desired output formats. We compare template-based programs (standard approach) and CFG-based variants, which are whitespace agnostic (comparison in \cref{app:whitespace_flexible}).
    \item \textbf{\llamacpp Grammars} \citep{llamacpp} We rely on \llamacpp's support for \lstinline|ebnf| grammars as an online parsing baseline (also representative of \cite{PoesiaP00SMG22} and \citet{geng2023grammar}).%
\end{itemize}

\begin{table*}
    \centering
    \footnotesize
    \caption{Impact on throughput of constrained decoding methods with different grammars, compared to unconstrained generation. We report the change in throughput compared to generating unconstrained output with the same model and inference backend. As CFG\textsuperscript{accel} we report \tool opportunistic masking or speculative decoding, depending on which is more effective.}
    \scalebox{1.0}{ 
    \begin{threeparttable}\
    \footnotesize
    \renewcommand{\arraystretch}{1.2}
    \centering
    \begin{tabular}{rcccccccccc}
    \toprule
     &  & 
        \makecell{\lstinline|llama.cpp|\textsuperscript{op} \\ \citet{llamacpp}} &
        \multicolumn{2}{c}{\makecell{\lstinline|guidance|\textsuperscript{\texttt{HF}} \\ \citet{guidance}}}&  
        \multicolumn{2}{c}{\makecell{\textbf{\tool}\textsuperscript{\texttt{HF}} \\ (Ours)}}
    \\ \cmidrule(lr){3-3} \cmidrule(lr){4-5} \cmidrule(lr){6-7}
    Grammar & Model & CFG & {\color{gray}Template\textsuperscript{$\downarrow$}} & CFG & CFG & CFG\textsuperscript{accel} \\
    \midrule
    \multirow{2}{*}{JSON (no schema)} &
            \emph{Mistral 7B} & 0.79$\times$ & {\color{gray}1.03$\times$} & 0.81$\times$ & 0.88$\times$ & \textbf{0.96$\times$} (opportunistic) \\
            & \emph{Llama 13B} & 0.83$\times$ & {\color{gray}1.03$\times$} & 0.86$\times$ & 0.95$\times$ & \textbf{1.12$\times$} (spec. $s=10$) \\
    \midrule
    \multirow{2}{*}{JSON (GSM8K schema, see \cref{app:structured_reasoning_outputs})} &
            \emph{Mistral 7B} & 0.80$\times$ & {\color{gray}0.77$\times$} & 0.59$\times$ & 0.99$\times$ & \textbf{1.77$\times$} (spec. $s=10$) \\
            & \emph{Llama 13B} & 0.86$\times$ & {\color{gray}0.94$\times$} & 0.74$\times$ & 0.99$\times$ & \textbf{1.66$\times$} (spec. $s=10$) \\
    \midrule
    \multirow{2}{*}{C Programming Language} &
            \emph{Mistral 7B} & 0.74$\times$ & - & - & 0.41$\times$ & \textbf{0.78$\times$} (opportunistic) \\
            & \emph{Llama 13B} & 0.81$\times$ & - & - & 0.54$\times$ & \textbf{0.85$\times$} (opportunistic) \\
    \midrule
    \multirow{2}{*}{XML (with schema)} &
            \emph{Mistral 7B} & 0.80$\times$ & - & - & 0.94$\times$ & \textbf{1.52$\times$} (spec. $s=10$) \\
            & \emph{Llama 13B} & 0.87$\times$ & - & - & 0.98$\times$ & \textbf{1.84$\times$} (spec. $s=10$) \\
    \midrule
    \multirow{2}{*}{Fixed Template} &
        \emph{Mistral 7B} & 0.55$\times$ & {\color{gray}1.95$\times$} & 0.92$\times$ & 0.97$\times$ & \textbf{1.30}$\times$ (spec. $s=10$) \\
      & \emph{Llama 13B} & 0.69$\times$ & {\color{gray}2.05$\times$} & 1.06$\times$ & 0.99$\times$ & \textbf{1.91}$\times$ (spec. $s=10$) \\
    \midrule
    \bottomrule
    \end{tabular}
    \begin{tablenotes}
        \footnotesize
        \item[\texttt{op}] \lstinline|llama.cpp| always runs with opportunistic masking, \textsuperscript{\texttt{HF}} Using the \lstinline|transformers| library as inference backend.
        \item[$\downarrow$] \guidance templates lead to significantly worse accuracy compared to CFGs (cf. \cref{tab:task_accuracy}), but we show them here for completeness.
    \end{tablenotes}    
    \label{tab:efficiency}
    \end{threeparttable}
    }
\end{table*}

\subsection{Task Accuracy}
\label{sec:task_accuracy_discussion}

We first evaluate the impact of different constrained decoding methods on downstream accuracy. For this, we use the \emph{GSM8K} benchmark for math reasoning and \emph{CoNLL-2003} for named-entity recognition. 

\paragraph{Setup} For both datasets, we prompt and constrain the models to generate a response in a given JSON format, instead of free-text reasoning (see \cref{app:structured_reasoning_outputs} for examples). Our prompts consist of 5 few-shot demonstrations from the training split, for which we manually construct the corresponding JSON response. We then compare the accuracy of the generated JSON responses, considering both the validity of the JSON format and the accuracy of the final response. We also measure perplexity of the generated output and the overhead over unconstrained generation in terms of throughput.

\paragraph{Results} As documented in \cref{tab:task_accuracy}, \tool achieves the best accuracy for all tasks, while also improving throughput well beyond unconstrained generation. In all cases, \tool's accuracy is the same or improved compared to unconstrained generation, indicating very low or no invasiveness. In contrast, with standard \guidance, we observe clear artifacts of invasiveness, as accuracy drops by up to $11\%$ points compared to unconstrained generation. And, while \guidance's inference optimizations do increase throughput (up to $2.02\times$), \tool accellerates inference even further (up to $2.71\times$), while also maintaining high accuracy. Implementing minimally invasive \guidance\textsuperscript{WS} programs with flexible whitespace and formatting restores some accuracy, but not fully, and also lowers throughput significantly to $\sim0.8\times$. \llamacpp's online parsing approach also does not appear to be fully non-invasive, although accuracy seems less impaired, while throughput is also consistently reduced to $\sim0.8\times$.

\subsection{Parameter Study}

Next, we investigate the key parameters of \tool.

\paragraph{Lookahead} To experiment with lookahead parameter $k$, we evaluate on GSM8K as before, but varying $k$. We report the results in \cref{tab:task_accuracy_k}. We observe that lower $k$ values impair performance significantly. Manual inspection shows that depending on $k$, the model is forced into different whitespacing behavior, as bridge tokens like \ttoken{\},} are unavailable, leading to irregularities. This can affect reasoning, e.g. \emph{Llama-2} is unable to produce object lists of length greater than 1, if relevant bridge tokens are missing. \tool with $k=\infty$, however, recovers and even slightly exceeds unconstrained accuracy, demonstrating minimal invasiveness.

\begin{table}
    \centering
    \caption{GSM8K task accuracy with different lookahead $k$.}
    \vspace{0.5em}
    \scalebox{0.8}{
    \begin{threeparttable}
    \centering
    \begin{tabular}{lll}
    \toprule
    \textbf{Configuration} & \emph{Mistral 7B} & \emph{Llama-2 13B} \\
    \midrule
            Unconstrained & 0.415 & 0.155 \\
            \tool ($k=0$) & 0.308 & 0.0 \\
            \tool ($k=1$) & 0.1 & 0.036 \\
            \tool ($k=\infty$) & \textbf{0.418} & \textbf{0.157} \\
    \midrule
    \bottomrule
    \end{tabular}
    \label{tab:task_accuracy_k}
    \end{threeparttable}
    }
\end{table}

\begin{figure}
    \centering
    \includegraphics[width=0.49\columnwidth]{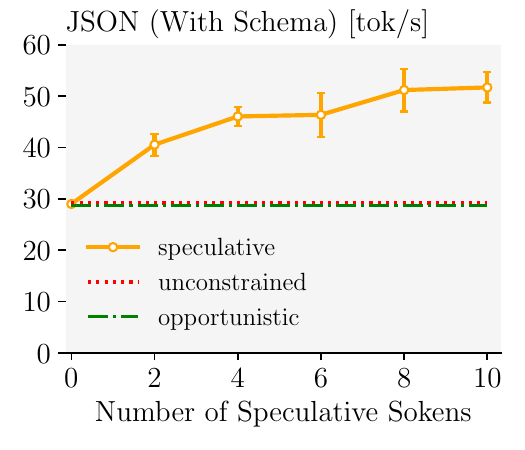}
    \hfill
    \includegraphics[width=0.49\columnwidth]{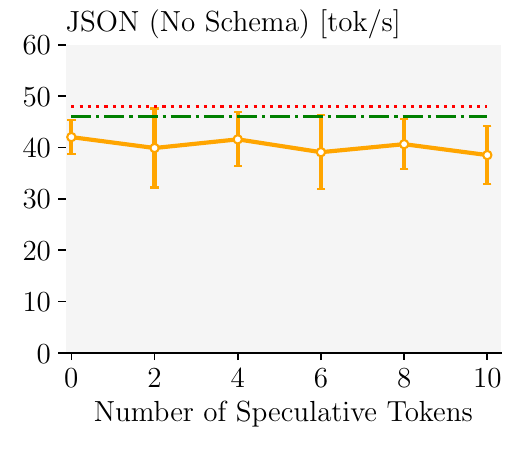}
    \caption{Impact of the number of speculative tokens $k$ on throughput (tokens per second) with \emph{Mistral 7B} and JSON generation with and without schema, using \tool with \lstinline|transformers| LLMs.}
    \label{fig:speculative_tokens}
\end{figure}

\paragraph{Speculation and Opportunistic Masking} We also experiment with the number of speculative tokens $s$ we propose at each decoding step, to speed up generation. For this, we compare the generation of schema-driven JSON and free-form JSON text with \emph{Mistral 7B}. We form priors on 10 randoms samples, and then measure mean performance across $100$ generated outputs per configuration, without updating counts. We report the results in \cref{fig:speculative_tokens}. 

We find speculative decoding with $s\in\{6,8,10\}$ to be particularly effective for schema-driven JSON generation, as it achieves a throughput of 1.7x compared to unconstrained generation. On free-form JSON output, speculation is not effective, and \tool opportunistic masking is preferable, incurring only a low $4 .21\%$ overhead. 

\subsection{Efficiency} Next, we examine throughput and efficiency of \tool. For this, we compare unconstrained generation thrroughput of each backend, and report the relative differences when running with constrained generation. We do not include \tool's precomputation time as part of the reported throughputs. We note that for the tested grammars, it ranges from 1-5s, with C being an outlier at around $20$s.

\paragraph{Grammars} We compare different constraining tasks, as shown in \cref{tab:efficiency}. For each task, we prepare a small set of relatively general inputs, prompting the model to generate a response in the general distribution of the given output format (details in \cref{app:grammars}). Where practical, we also implement \guidance programs, using their custom CFG-like syntax.

\paragraph{Setup} We run 100 repetitions per configuration. In each, we sample one of 5 different prompts per workload, and sample output of up to $128$ tokens from the model, using a temperature value of 1.0. This way, we ensure that the model produces in-distribution output, but that it still exhibits diversity. Before measuring, we run $10$ repetitions of warmup, allowing our speculative mechanisms to form a prior. After that, the learned priors remain fixed.

\paragraph{Results} As shown in \cref{tab:efficiency}, \tool is highly effective and clearly reduces the compuational load of constraining at inference time. \tool outcompetes both \guidance and \lstinline|llama.cpp|'s online parsing approach significantly. For grammars with predictable structure (e.g. schema-driven formats), speculative decoding is particularly effective, leading to up to $77\%$ higher throughput over unconstrained generation, while remaining minimally invasive.

C code generation induces the most overhead, which can be explained by the fact that the C grammar is the most complex of the tested workloads. Here, speculative decoding does not bring any benefits, as the C code is to hard to predict using our simple count-based model. However, by relying on \tool's opportunistic masking mode, we still outcompete \lstinline|llama.cpp|, running at $0.78\times$ vs. $0.74\times$.

\section{Conclusion} \label{sec:conclusion}
We have shown the need for minimally invasive, highly-efficient constrained decoding methods. As first instantiation of this for grammars, we presented \tool, which leverages precomputation, speculative decoding and opportunistic masking, to implement
minimally invasive constraining (no accuracy loss), often overhead-free or even faster generation, and thus, high-throughput inference.

\section*{Impact Statement}
This paper presents work whose goal is to advance the field of Machine Learning. There are many potential societal consequences of our work, none which we feel must be specifically highlighted here.

\section*{Acknowledgements}
This work has received funding from the Swiss State Secretariat for Education, Research and Innovation (SERI).
This work has been done as part of the SERI grant SAFEAI (Certified Safe, Fair and Robust Artificial Intelligence, contract no. MB22.00088, SERI-funded ERC Consolidator Grant).

\message{^^JLASTBODYPAGE \thepage^^J}

\clearpage
\bibliography{references}
\bibliographystyle{icml2024}

\message{^^JLASTREFERENCESPAGE \thepage^^J}

\ifbool{includeappendix}{%
	\clearpage
	\appendix
	
\section{Whitespace-Flexible \guidance Programs}
\label{app:whitespace_flexible}

In our experiments, we differentiate standard \guidance programs based on templates and whitespace-flexible \guidance\textsuperscript{WS} programs.

To demonstrate, consider the following example of a simple template-based \guidance program:

\begin{lstlisting}[escapeinside={(((*}{*)))}, basicstyle=\ttfamily\scriptsize, breaklines=true, caption={A standard JSON \guidance program.}]
f"""{{
      "id": {gen('id', regex='[1-9][0-9]*')},
      "description": "A nimble fighter",
      "name": "{gen('name', stop='"')}",
      "age": {gen('age', regex='[1-9][0-9]*')},
      "armor": "{select(['leather', 'chainmail', 'plate'])}",
      "weapon": "{select(['sword', 'axe', 'bow'])}",
      "class": "{gen('class', stop='"')}",
      "mantra": "{gen('mantra', stop='"')}",
      "strength": {gen('strength', regex='[1-9][0-9]*')},
      "items": [ "{gen('item', stop='"')}", "{gen('item', stop='"')}", "{gen('item', stop='"')}" ],
}}"""
\end{lstlisting}

Here, we provide a fixed high-level template, with respect to whitespace and formatting. Only the values of the fields are generated by the LLM.

In contrast, a whitespace-flexible \guidance\textsuperscript{WS} program for the same task would look as follows:

\begin{lstlisting}[escapeinside={(((*}{*)))}, basicstyle=\ttfamily\scriptsize, breaklines=true, caption={A whitespace-flexible JSON \guidance program.}]
nl = "\n"
WS = token_limit(zero_or_more(select([' ', nl])), 16)

f"""{{{WS}"id"{WS}:{WS}{gen('id', regex='[1-9][0-9]*')}{WS},{WS}"description":{WS}"A nimble fighter"{WS},{WS}"name"{WS}:{WS}"{gen('name', stop='"')}"{WS},{WS}"age"{WS}:{WS}{gen('age', regex='[1-9][0-9]*')}{WS},{WS}"armor"{WS}:{WS}"{select(['leather', 'chainmail', 'plate'])}"{WS},{WS}"weapon"{WS}:{WS}"{select(['sword', 'axe', 'bow'])}"{WS},{WS}"class":{WS}"{gen('class', stop='"')}"{WS},{WS}"mantra"{WS}:{WS}"{gen('mantra', stop='"')}"{WS},{WS}"strength"{WS}:{WS}{gen('strength', regex='[1-9][0-9]*')}{WS},{WS}"items":{WS}[{WS}"{gen('item', stop='"')}"{WS},{WS}"{gen('item', stop='"')}"{WS},{WS}"{gen('item', stop='"')}"{WS}]{WS},{WS}}}"""
\end{lstlisting}

As shown in the snippet, all explicit templated whitespace is replaced by a \lstinline|{WS}| token, using the \lstinline{zero_or_more} operator. Using this approach, all whitespace is now also generated by the LLM, allowing for more flexible formatting of the output, and less explicit constraints on the LLM.

Our experiments in \cref{sec:eval} demonstrate that the whitespace-flexible formulation leads to higher task accuracy but also significantly higher inference time. This is because the program leaves more freedom to the LLM on how to concretely generate the output. At the same time however, inference becomes less efficient, as the LLM now also has to generate all whitespace tokens explicitly and the \guidance runtime cannot skip over as many tokens as before.

\section{Model-Based Retokenization}
\label{app:retokenization}

To demonstrate differences between template-based and unconstrained generation, we consider the task of \emph{naturalizing} a given text under a model-preferred tokenization. This is the process of converting text to the tokenization a model would have chosen to represent the same text during generation, when previously conditioned on some prompt. More specifically, given a target text $s$, a tokenized prompt $x$, and a model $f$ with vocabulary \voc, we re-encode $t$ using tokens from $\voc$, such that we greedily maximize the sequence likelihood assigned by $f$. 

We refer to this process as \emph{retokenization}. We provide the procedure for this in \cref{alg:retokenization}. By greedily choosing the highest likelihood token that aligns with the target text, we obtain a tokenization of $s$, that is consistent with the model's preference, when forced to generate $s$ from $x$. This corresponds to applying $\argmax$ decoding to $f$, where the token distribution of $f$ is always masked such that it produces the target text $s$. Put differently, if a model was to generate $s$ when conditioned on $x$, it would have produced the token sequence $o$ under $\argmax$ decoding.

While retokenization allows us to recover the model-preferred tokenization of a given text, template-based constrained generation methods cannot benefit from this, as its computational overhead is equivalent to the cost of generating all templated tokens from scratch, thereby negating the benefits of using a template-based approach in the first place.

\begin{algorithm}[t]
    \caption{Model-Based Retokenization} \label{alg:retokenization}
    \begin{algorithmic}[1]
    \REQUIRE LLM $f$, Prompt $x$, Target Text $s$
    \ENSURE $f$-preferred tokenization of $s$
    \STATE $o \leftarrow []$
    \WHILE{$s \neq \emptyset$}
    \STATE $\vv \leftarrow f(x + o)$ \COMMENT{compute logits}
    \STATE $t \leftarrow \argmax\left\{\vv[t] \mid t \in \voc \wedge t\text{ prefix of } s\right\}$
    \STATE $o.\ttt{append}(t)$
    \STATE $s \leftarrow s\text{[|o|:]}$ \COMMENT{remove prefix $t$ from $s$}
    \ENDWHILE
    \STATE \textbf{return} $o$
    \end{algorithmic}
\end{algorithm}

\section{Grammars And Prompts} \label{app:grammars}

Below we include the grammars and prompts used for each constraining task from our experiments in Section~\ref{sec:eval}:

\begin{lstlisting}[escapeinside={(((*}{*)))}, basicstyle=\ttfamily\scriptsize, breaklines=true, caption={Basic JSON Grammar}]
root   ::= object
value  ::= object | array | string | number | 
           ("true" | "false" | "null") ws

object ::=
  "{" ws (
            string ":" ws value
    ("," ws string ":" ws value)*
  )? "}" ws

array  ::=
  "[" ws (
            value
    ("," ws value)*
  )? "]" ws

string ::=
    "\"" (
    [^"\\] |
    "\ \" (["\ \/bfnrt] | 
    "u" [0-9a-fA-F] 
    [0-9a-fA-F] 
    [0-9a-fA-F] 
    [0-9a-fA-F]) # escapes
  )* "\"" ws

number ::= ("-"? ([0-9] | 
             [1-9] [0-9]*)) 
           ("." [0-9]+)? ([eE] [-+]? [0-9]+)? ws

ws ::= ([ \ \ t\ \ n] ws)?

# Prompts used for generation
"A JSON file describing a person:"
"A JSON file of a person John Smith:"
"A JSON file of a person John Smith with friends"
"JSON of a person Jane Doe with friends"
"A JSON person:"
\end{lstlisting}

\begin{lstlisting}[escapeinside={(((*}{*)))}, basicstyle=\ttfamily\scriptsize, breaklines=true, caption={Guided Math Reasoning Grammar (for GSM8K)}]
root   ::= object
value  ::= object | array | string | number | ("true" | "false" | "null") ws

object ::=
    ws "{" ws (
        "\"thoughts\"" ":" ws "[" ws thought (ws "," ws thought)* "]" ws "," ws
        "\"answer\"" ":" ws number ws
    ) "}" ws

thought ::=
    "{" ws (
        "\"step\"" ":" ws string "," ws
        "\"calculation\"" ":" ws string "," ws
        "\"result\"" ":" ws number
    ) "}" ws

array  ::=
    "[" ws (
            value
    ("," ws value)*
    )? "]" ws

string ::=
    "\"" (
    [^"\\] |
    "\\" (["\\/bfnrt] | "u" [0-9a-fA-F] [0-9a-fA-F] [0-9a-fA-F] [0-9a-fA-F]) # escapes
    )* "\"" ws

number ::= ("-"? ([0-9] | [1-9] [0-9]*)) ("." [0-9]+)? ([eE] [-+]? [0-9]+)? ws

# Optional space: by convention, applied in this grammar after literal chars when allowed
ws ::= ([ \t\n] ws)?

# Prompts used for generation
We use 5-shot prompts for questions from GSM8K's test split as prompt (cf. Task Accuracy Experiments).
\end{lstlisting}

\begin{lstlisting}[escapeinside={(((*}{*)))}, basicstyle=\ttfamily\scriptsize, breaklines=true, caption={Simple C Program Grammar and Prompts}]
root ::= (declaration)*

declaration ::= dataType identifier ws "(" ws parameter? ws ")" ws "{" ws statement* "}"

dataType  ::= "int" ws | "float" ws | "char" ws
identifier ::= [a-zA-Z_] [a-zA-Z_0-9]*

parameter ::= dataType identifier

statement ::=
    ( dataType identifier ws "=" ws expression ";" ws ) |
    ( ( dataType identifier ws "[" ws expression ws "]" ws ( "=" ws expression )? ";" ws ) ) |
    ( identifier ws "=" ws expression ";" ws ) |
    ( identifier ws "(" argList? ")" ";" ws) |
    ( "return" ws expression ";" ws ) | 
    ( "while" "(" condition ")" ws "{" statement* "}" ) |
    ( "for" "(" forInit ";" ws condition ";" ws forUpdate ")" "{" statement* "}" ws ) |
    ( "if" "(" condition ")" "{" statement* "}" ("else" "{" statement* "}")? ws ) |
    ( singleLineComment ws ) |
    ( multiLineComment ws )

forInit ::= dataType identifier ws "=" ws expression | identifier ws "=" ws expression
forUpdate ::= identifier ws "=" ws expression

condition ::= expression relationOperator expression
relationOperator ::= ("<=" | "<" | "==" | "!=" | ">=" | ">")

expression ::= term (("+" | "-") term)*
term ::= factor(("*" | "/") factor)*

string ::=
    "\"" (
    [^"\\] |
    "\\" (["\\/bfnrt] | "u" [0-9a-fA-F] [0-9a-fA-F] [0-9a-fA-F] [0-9a-fA-F]) # escapes
    )* "\"" ws

factor ::= identifier | number | unaryTerm | funcCall | parenExpression | subscript | string
unaryTerm ::= "-" factor
funcCall ::= identifier "(" argList? ")"
parenExpression ::= "(" ws expression ws ")"
subscript ::= identifier "[" ws expression ws "]"

argList ::= expression ("," ws expression)*

number ::= [0-9]+

singleLineComment ::= "//" [^\n]* "\n"
multiLineComment ::= "/*" ( [^*] | ("*" [^/]) )* "*/"

ws ::= ([ \t\n]*)

# prompts used for generation
"A C program that prints \"Hello, world!\":\n```c\n"
"A C main function that iterates over an array of integers and prints each one:\n```c\n"
"A C program that prints the sum of two integers:\n```c\n"
"The following is a program that finds the sum of two integers in C:\n```c\n"
"A C program that fills an array with the numbers 0 to 9 and prints them:\n```c\n"
"A C implementation of a simple bubble sort:\n```c\n"
\end{lstlisting}

\begin{lstlisting}[escapeinside={(((*}{*)))}, basicstyle=\ttfamily\scriptsize, breaklines=true, caption={XML (with schema) Grammar and Prompts}]
root ::= person

person ::= ( "<person>" ( ws personattributes ) "</person>" )
personattributes ::= nameattribute ageattribute jobattribute friends?

nameattribute ::= "<name>" NAME "</name>" ws
ageattribute ::= "<age>" NUMBER "</age>" ws
jobattribute ::= "<job>" ws jobinfo "</job>" ws
friends ::= "<friends>" ws person+ ws "</friends>" ws

jobinfo ::= jobtitle jobsalary
jobtitle ::= "<title>" NAME "</title>" ws
jobsalary ::= "<salary>" NUMBER "</salary>" ws

NAME ::= ( [^<] )+
NUMBER ::= ( [^<] )+

# Optional space: by convention, applied in this grammar after literal chars when allowed
ws ::= ([ \t\n] ws)?

# prompts used for generation
"An XML file describing a person:"
"An XML file of a person John Smith:"
"An XML file of a person John Smith with friends"
"XML of a person Jane Doe with friends"
"An XML person:"
\end{lstlisting}

\begin{lstlisting}[escapeinside={(((*}{*)))}, basicstyle=\ttfamily\scriptsize, breaklines=true, caption={Fixed Template Grammar and Prompts}]
start: dict

dict: "{" content "}"

content: id_pair "," description_pair "," name_pair "," age_pair "," armor_pair "," weapon_pair "," class_pair "," mantra_pair "," strength_pair "," items_pair

id_pair: "\"id\"" ":" NUMBER
description_pair: "\"description\"" ":" "\"A nimble fighter\""
name_pair: "\"name\"" ":" STRING
age_pair: "\"age\"" ":" NUMBER
armor_pair: "\"armor\"" ":" (("\"leather\"") | ("\"chainmail\"") | ("\"plate\""))
weapon_pair: "\"weapon\"" ":" (("\"sword\"") | ("\"axe\"") | ("\"bow\""))
class_pair: "\"class\"" ":" STRING
mantra_pair: "\"mantra\"" ":" STRING
strength_pair: "\"strength\"" ":" NUMBER
items_pair: "\"items\"" ":" "[" item "," item "," item "]"

item: STRING

STRING: /"[^\n\r"]+"/
NUMBER: /[0-9]+/

WS: /[ \t\n]+/

# prompts used for generation
"The following is a character profile for an RPG game in JSON format.\n```json\n",
"A character profile for an RPG game:\n```json\n",
"A character profile for an RPG game in JSON format:\n```json\n",
"A character that is a level 5 human fighter with 10 strength, 10 dexterity, 10 constitution, 10 intelligence, 10 wisdom, and 10 charisma:\n```json\n",
"JSON specifying a character that is a level 5 dwarf fighter from a game:\n```json\n"
\end{lstlisting}

\section{Structured Reasoning Outputs}
\label{app:structured_reasoning_outputs}

In our experiments, we evaluate task accuracy on GSM8K and CoNLL2003. For these tasks, prompted and constrained model output looks as follows:

\begin{lstlisting}[escapeinside={(((*}{*)))}, basicstyle=\ttfamily\scriptsize, breaklines=true, caption={Structured Reasoning Output for GSM8K}]
{
    "thoughts": [
        {
            "step": "Find the distance between the first and second stops",
            "calculation": "60 - 20 - 15",
            "result": 25
        },
        {
            "step": "Find the distance between the first and second stops",
            "calculation": "25 + 15",
            "result": 40
        }
    ],
    "answer": 40
}
\end{lstlisting}

\begin{lstlisting}[escapeinside={(((*}{*)))}, basicstyle=\ttfamily\scriptsize, breaklines=true, caption={Structured Reasoning Output for CoNLL2003}]
{
    "tokens": [
        {
            "token": "Nadim",
            "tag": "B-PER"
        },
        {
            "token": "Ladki",
            "tag": "I-PER"
        }
    ]
}
\end{lstlisting}

In practice, such outputs greatly facilitate downstream processing of LLM outputs, as they are already in a structured format and can be easily parsed.

\paragraph{Few-Shot Demonstrations} For few-shot demonstrations, we alternate between questions and answers using a simple \lstinline|Q: ... \n A: ... \n ...| format.

}{}

\message{^^JLASTPAGE \thepage^^J}

\end{document}